\newcommand{\CLASSINPUTtoptextmargin}{21.2mm}
\newcommand{\CLASSINPUTbottomtextmargin}{15.2mm}
\newcommand{\CLASSINPUTinnersidemargin}{16.9mm}
\newcommand{\CLASSINPUToutersidemargin}{16.9mm}

\documentclass[letterpaper, 10 pt, conference]{IEEEtran} 
\IEEEoverridecommandlockouts
\usepackage{graphicx} 
\usepackage{mathtools} 
\usepackage{siunitx}
\usepackage{cite}
\usepackage{amsmath,amsfonts,amssymb,amsthm}
\usepackage{mathtools}
\usepackage{commath}
\usepackage{pdfpages}
\usepackage{svg}
\usepackage{multirow}
\usepackage{algorithmic}
\usepackage{gensymb}
\usepackage{graphicx}
\usepackage{textcomp}
\usepackage[left=.75in, right=.75in, top=.75in, bottom=.8in]{geometry}
\usepackage{amsmath}
\usepackage{dblfloatfix} 
\usepackage{xcolor}
\usepackage{comment}
\usepackage{float}

\usepackage{fancyhdr}
\fancypagestyle{withfooter}{
  
  \fancyfoot[C]{\footnotesize Accepted to the IEEE ICRA Workshop on Field Robotics 2024}
}

\def\BibTeX{{\rm B\kern-.05em{\sc i\kern-.025em b}\kern-.08em
    T\kern-.1667em\lower.7ex\hbox{E}\kern-.125emX}}
    
\newgeometry{left=.75in, right=.75in, top=1in, bottom=.8in}

\usepackage{color}
\usepackage{colortbl}
\definecolor{Gray}{gray}{0.9}
\usepackage[colorinlistoftodos]{todonotes}

\begin{document}

\title{ReachBot Field Tests in a Mojave Desert Lava Tube as a Martian Analog}

\author{\IEEEauthorblockN{Tony G. Chen$^1$, Julia Di$^1$, Stephanie Newdick$^2$, Mathieu Lap\^{o}tre$^{3}$, Marco Pavone$^2$, Mark R. Cutkosky$^1$}

\thanks{$^1$Tony Chen, Julia Di, and Mark Cutkosky are with the Department of Mechanical Engineering, Stanford University, Stanford, CA 94305. \{{\tt agchen, juliadi, cutkosky}\} {\tt@stanford.edu}.}

\thanks{$^2$Stephanie Newdick, Marco Pavone are with the Department of Aeronautics and Astronautics, Stanford University, Stanford, CA 94305. \{{\tt snewdick, pavone}\} {\tt@stanford.edu}.}

\thanks{$^2$Mathieu Lap\^{o}tre is with the Department of Earth and Planetary Sciences, Stanford University, Stanford, CA 94305. \{{\tt mlapotre}\} {\tt@stanford.edu}.}

}

\maketitle

\begin{abstract}
ReachBot is a robot concept for the planetary exploration of caves and lava tubes, which are often inaccessible with traditional robot locomotion methods. It uses extendable booms as appendages, with grippers mounted at the end, to grasp irregular rock surfaces and traverse these difficult terrains. We have built a partial ReachBot prototype consisting of a single boom and gripper, mounted on a tripod. We present the details on the design and field test of this partial ReachBot prototype in a lava tube in the Mojave Desert. The technical requirements of the field testing, implementation details, and grasp performance results are discussed. The planning and preparation of the field test and lessons learned are also given.

\end{abstract}

\section{Introduction}

There is a growing interest in space operations that require robots capable of mobile manipulation under a variety of gravity
regimes and terrain geometries. NASA's interest in exploring caves, cliffs, and other rocky terrain on Mars~\cite{NRC2011,LapotreORourkeEtAl2020} motivates the need for robots that combine sparse-anchoring mobility with high-wrench manipulation, a combination that existing solutions do not satisfy. Small robots are typically restricted to limited reach and wrench capability. Conversely, large robots, particularly rigid-link articulated-arm robots, are hampered by mass and complexity, which scale poorly with increased reach.

\begin{figure}[]
\centering
\includegraphics[width=.49\textwidth]{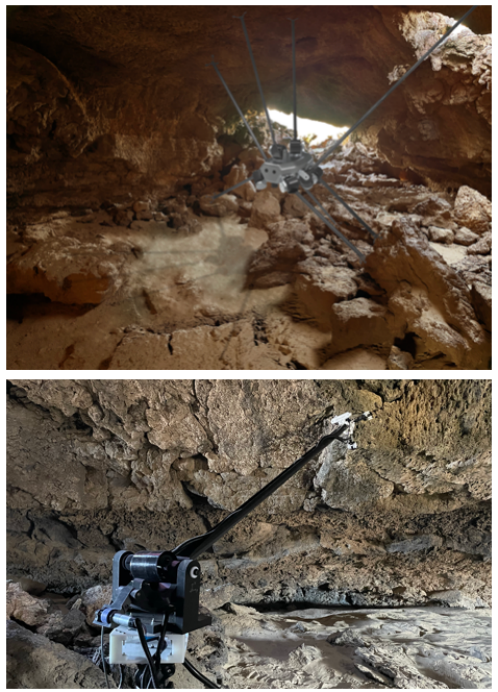}
\caption{ReachBot is a robotic concept that uses extending booms as prismatic limbs to navigate difficult terrain. Shown in a rendering (top) is the full ReachBot configuration overlaid on a photograph of the field test site in the Lavic Lake volcanic field in the Mojave Desert. During this field trial, a single-boom prototype (bottom) was tested to demonstrate a boom deployer, perception system, and microspine gripper.}
\label{fig:fig1}
\end{figure}

ReachBot is a robot concept that locomotes by grasping rock features with multiple appendages, traversing environments that require climbing even when anchor points are sparse (Fig.~\ref{fig:fig1}). 
Using lightweight, extendable booms as appendages, ReachBot achieves a form factor that has a small body with very long limbs.
Extendable booms have been developed for space applications such as antenna structures \cite{Fernandez2017, FootdaleMurphey2014, SpenceWhiteEtAl2018} because they are light and compact when rolled up, but are strong---especially in tension---when deployed and capable of extending many times the span of the robot body. 
For ReachBot, the tips of the booms are equipped with pivoting wrists and grippers that use arrays of microspines to grasp rocky surfaces. This arrangement allows ReachBot to move by manipulating its body with respect to the terrain. 

In previous publications, the ReachBot concept has been introduced, with analysis given on planar motion planning and control \cite{SchneiderBylardEtAl2022, ChenMillerEtAl2022, NewdickChenEtAl2023, NewdickOngoleEtAl2023}. This analysis has been recently extended to fully three-dimensional scenarios and reported in \cite{chen2024reach}.  In this paper we expound upon the field experience and lessons learned from a deployment of ReachBot technology in a Mojave Desert lava tube.

\section{ReachBot Prototype}

We have developed a partial ReachBot prototype for a field test in a relevant lava tube environment, which consists of a single boom deployer mounted on a shoulder joint that allowed it to pan and tilt. The gripper is mounted at the end of the boom along with an Intel Realsense d455 camera. The entire assembly is mounted on a tripod, representing the body of ReachBot, and is controlled via teleoperation. The overall assembly is shown in Fig.~\ref{fig:fig1} (bottom).

\subsection{Grippers}
For ReachBot's locomotion, it needs to compute each grasps' expected maximum magnitude of the pulling force that it is possible to withstand before any grasp failure, along with the direction of this force for a given surface geometry. This calculation has a stochastic nature since it is impossible to have knowledge about the individual rock asperities, which are the features that the microspines on the gripper engage with. We deal with this problem by (i) developing a grasp model, which allows us to compute the contact forces on the individual fingers, and (ii) by approximating the distribution (mean and variance) of the pulling force magnitude through Monte Carlo sampling.


The resulting gripper is a lightweight underactuated gripper specialized for secure grasps on irregular convex surfaces. It has a single, non-backdrivable motor controlling a closing tendon and an opening tendon through a load-sharing whiffetree mechanism. There are linearly-independent spines at the distal phalange of each finger with redundant degrees-of-freedom to conform to the highly irregular surfaces.

Many parts of the gripper were manufactured using 3D printing technology. The spine tiles were printed with Stratasys Objet VeroWhite to achieve the dimensional accuracy required. The fingers were printed with HP Multi Jet Fusion (MJF) Nylon 12 PA. The rest of the gripper was printed on a Formlab3+ with Rigid 4k material for rigidity. The connections between parts were accomplished by press-fitting brass screw-to-expand inserts. The motor (Pololu 1000:1 Micro Metal Gearmotor) underwent a further 30:1 gear reduction through a wormgear assembly (Misumi SUW0.5-R1 and G50B20+R1) and controlled the loading tendon (Twinline braided Vectran 125) and opening tendon (Power Pro Spectra 40\,lbs Fishing Line (178\,N)).

\subsection{Vision System}
Grasp strength is strongly dependent on surface profile, but it is costly to deploy and reposition a gripper towards a new site. Because of the commitment required to extend a boom, the prototype ReachBot used a two-stage surface scanning process which involved scanning the surface from afar to determine a general region of interest, and then re-scanning the surface from a closer distance to determine the exact grasp location.

To achieve this two-step scanning approach, an Intel Realsense d455 RGBD camera was rigidly mounted to the distal end of the boom, co-located with the gripper wrist. The camera provided data from a range of $0.6$ - $\SI{6}{\meter}$. The camera streamed images to an offboard laptop which processed depth images into pointclouds for grasp site determination.

While the boom was stowed, a pointcloud was processed and spheres of best fit were approximated onto the surface using the M-estimator SAmple Consensus (MSAC) algorithm in MatLab. Then, the boom was extended toward the region of the surface with the closest spherical features. Before committing to the grasp site, another algorithm was run to determine the local normal from the \textit{k}-nearest neighbors and compare against a range of acceptable values.

\begin{figure}[]
\centering
\includegraphics[width=.48\textwidth]{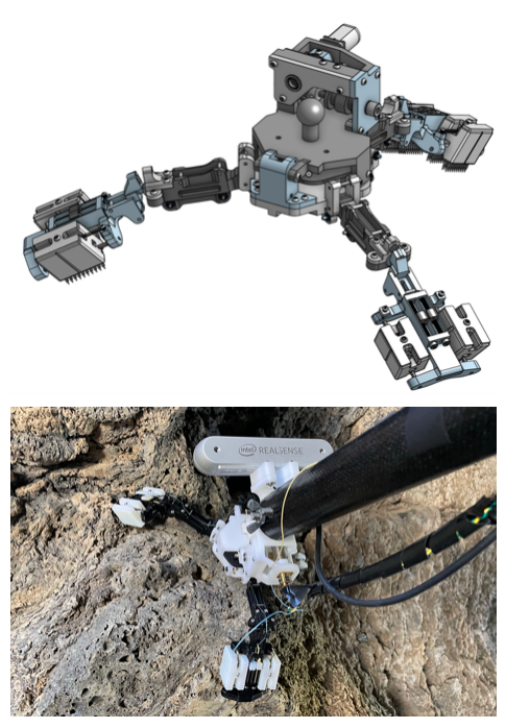}
\caption{A custom microspine gripper was developed and field tested. Shown is a CAD rendering (top) of the microspine gripper. Also shown is a photograph (bottom) from the field test of the gripper successfully grasping a rock while mounted at the end of an extended boom.}
\label{fig:fig2}
\end{figure}

\subsection{Prototype Assembly}
The deployer was custom-built with three different actuators to provide the necessary functionality of a ReachBot shoulder joint. The deploying mechanism (3D printed, Creality Ender 3 PLA) was a friction-drive wheel powered by a DC motor (Pololu 75:1 Metal Gearmotor). This drive wheel was pressed against the boom coil by compression springs. An additional DC motor (Pololu 75:1 Metal Gearmotor) controlled the elevation of the boom through timing belts and pulleys. The entire assembly was mounted on a turntable (McMaster Carr 6031K16) powered by a servo (Zoskay 35kg) that controlled the pan angle.  

The deployer and gripper were controlled by a single Arduino Mega with manual input from a joystick, and serial commands through USB from a laptop for the gripper. The pan-angle servo was driven using PWM signals from the Arduino Mega. All DC motors were powered by a 12 V power supply and the servo motor was powered by a separate 6 V power supply.

\section{Test Site Considerations}

As noted earlier, planetary caves and lava tubes are among the most promising geological and astrobiological targets in the solar system \cite{blank2018planetary,titus2021roadmap,wynne2022planetary,wynne2022fundamental}. Given that martian lava tubes are of particular interest \cite{LeveilleDatta2010, phillips2020mars} it was decided that a field test of ReachBot technology should be conducted in a reasonable analog on Earth. 

To determine the test site location, we conducted an initial survey of the Mojave Desert Lavic Lake volcanic field to identify candidate locations for conducting the field test. The Lavic Lake volcanic field consists of Quaternary (likely late Pleistocene \cite{phillips2003cosmogenic}) basaltic pahoehoe and aa lava flows \cite{wise1966geologic} and has a long history of serving as a planetary analog \cite{greeley1988relationship,arvidson1998rocky}. The presence of both pahoehoe and aa flow textures ensures that the ReachBot prototype is subjected to a wide range of surface roughnesses, similar to what could be encountered on Mars. 

\begin{figure}[h]
\centering
\includegraphics[width=.48\textwidth]{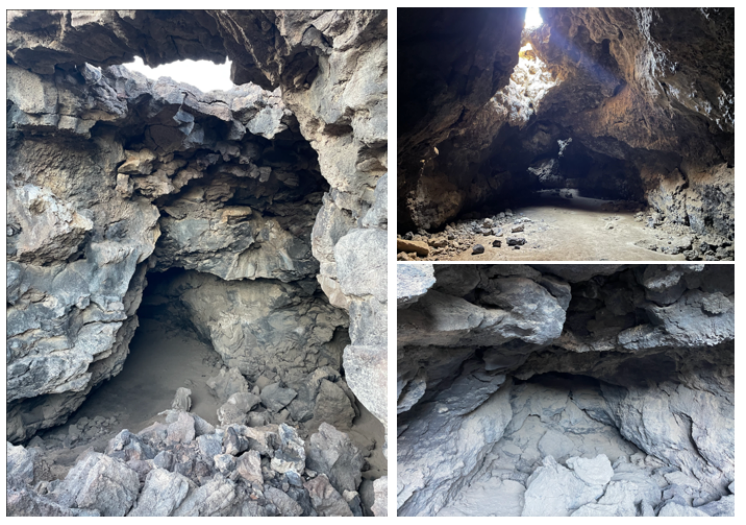}
\caption{Shown are photographs of a few candidate lava tubes that were considered during the initial survey of the Mojave Desert Lavic Lake volcanic field.}
\label{fig:caves}
\end{figure}

During this initial survey, we identified four local lava tubes that were satisfactory analogs, with some examples shown in Fig. \ref{fig:caves}. To determine the final testing location, we considered the following practical features: entrance size, ease of access to the cave, distance from the closest road, and safety to human operators. After careful consideration of the criteria, the field tests were conducted in an unnamed lava tube near Pisgah Crater, within the Lavic Lake volcanic field in the Mojave Desert, California (34\degree 44'59.0" N 116\degree22'03.1" W). 

\section{Field Test Report}

\subsection{Experimental setup}
To support the testing, a central control station consisting of a folding table, a laptop, two electrical power supplies, and a portable power station (Jackery Explorer 300) is set up inside the cave. The partial ReachBot prototype is mounted on a heavy duty tripod, placed approximately two meters from the potential grasping sites, in a central location in the cave. A simple gravity off-load system is rigged through a skylight entrance; Martian gravity is a third of that of the Earth, so a gravity off-load system is reasonable compensation for field testing purposes. The main objective of the field test was (i) a system-level demonstration of the partial ReachBot system from perception of the rock surface to a final grasp, and (ii) observation and data collection on the performance of the microspine gripper grasping natural, highly irregular rock surfaces. 

\subsection{Results}

\begin{figure*}[]
\includegraphics[width=1.0\textwidth]{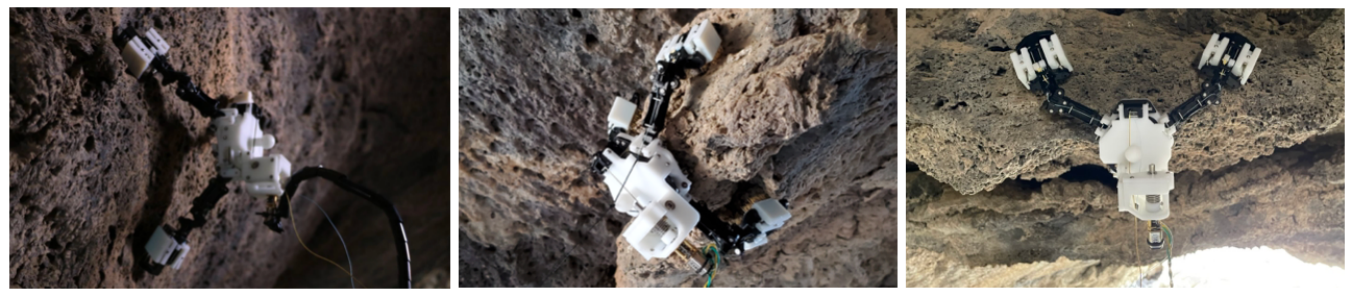}
\caption{In the field test, the microspine gripper was capable of grasping a variety of different rock surfaces, ranging from hemispherical (left) to ledge-like features (middle, right). These results provide confidence that the microspine gripper will successfully grasp highly irregular lavic rock surfaces.}
\label{fig:gripperrock}
\end{figure*}

The single-arm prototype was able to target, deploy and grasp many target sites within the lava tube.
For the perception system, the field test confirmed the advantages of a two-step perception strategy, starting with a remote scan with the RGBD camera. The remote scan, conducted under teleoperation, identified areas that were expected to yield fruitful grasp locations. Figure \ref{fig:gripperrock} shows grasps on features with approximately spherical (radially symmetric) and cylindrical geometry, respectively. Such features were detectable using the perception system. 

For the gripper, a wide range of candidate grasp locations were successfully tested using the real gripper under its own power, as shown in Fig. \ref{fig:gripperrock}. In these images, the boom has been removed for measuring pull-off forces with a Mark-10 (model M4-50 with $\SI{0.1}{\newton}$ resolution) force gauge. The gripper is loaded in the direction of the boom.
In all cases, the grasps achieved at least $\SI{34.2}{\newton}$ (larger forces were not applied to avoid breaking the 3D-printed plastic components of the prototype during field testing). 

Additional gripper pull-off tests with the Mark-10 were conducted on a large lava rock sample acquired at the testing site. In these tests, the gripper was placed on a surface of the rock and then pulled until the grasp failed, with the maximum forces and pulling angle recorded. 

There are many lessons learned from the field test conducted in the Mojave Desert, which will provide insights into future expeditions for robots in lava tubes. The first is the ease of access to the lava tube, which is usually in areas that are very difficult, if not impossible, to traverse by vehicles. These lava fields provide many challenges in transporting robots and testing equipment to the desired location. The second is the accumulation of dust on the robot due to iron-rich minerals in the lava rocks and the surrounding sand. These fine dust particles are attracted to robot components that are magnetic, such as motors. Lastly, the only natural light sources in these lava tubes are from the entrances and skylights, if one exists. Adequate lighting equipment should be brought along for field testing, recording, and photography purposes.




\section{Acknowledgements}
This work was supported by the NASA Innovative Advanced Concepts (NIAC) program. S. Newdick and J. Di were supported by NASA Space Technology Graduate Research Opportunities (NSTGRO). T.G. Chen was supported by the  NSF Graduate Research Fellowships Program (GRFP).

\bibliographystyle{IEEEtran}
\bibliography{mybib.bib,ASL_papers}

\end{document}